# Lp—A logic for Statistical Information

Fahiem Bacchus[*]
Department of Computer Science
University of Waterloo
Waterloo, Ontario, Canada
N2L-3G1
fbacchus@watdragon.waterloo.edu

## 1 Introduction

This extended abstract presents a logic, called **Lp**, that is capable of representing and reasoning with a wide variety of both qualitative and quantitative statistical information. The advantage of this logical formalism is that it offers a declarative representation of statistical knowledge. Knowledge represented in this manner can be used for a variety of reasoning tasks.

The logic differs from previous work in probability logics in that it uses a probability distribution over the domain of discourse, whereas most previous work (e.g., Nilsson [1], Scott et al. [2], Gaifman [3], Fagin et al. [4]) has investigated the attachment of probabilities to the sentences of the logic (also, see Halpern [5] for further discussion of the differences).

The logic **Lp** possesses some further important features. First, **Lp** is a superset of first order logic, hence it can represent ordinary logical assertions. This means that **Lp** provides a mechanism for integrating statistical information and reasoning about uncertainty into systems based solely on logic. Second, **Lp** possesses transparent semantics, based on sets and probabilities of those sets. Hence, knowledge represented in **Lp** can be understood in terms of the simple primitive concepts of sets and probabilities. And finally, the there is a sound proof theory which has wide coverage (the proof theory is complete for certain classes of models). The proof theory captures a sufficient range of valid inferences to subsume most previous probabilistic uncertainty reasoning systems. For example, the constraints generated by Nilsson's probabilistic entailment [1] can be generated by the proof theory, and the Bayesian inference underlying belief nets [6] can be performed. In addition, the proof theory integrates quantitative and qualitative reasoning as well as statistical and logical reasoning.

In the next section we briefly examine previous work in probability logics, comparing it to **Lp**. Then we present some of the varieties of statistical information that **Lp** is capable of expressing. After this we present, briefly, the syntax, semantics, and proof theory of the logic. We conclude with a few examples of knowledge representation and reasoning in **Lp**, pointing out the advantages of the declarative representation offered by **Lp**. We close with a brief discussion of probabilities as degrees of belief, indicating how such probabilities can be generated from statistical knowledge encoded in **Lp**.

## 2 Other Probability Logics

Previous work in probability logic has investigated the attachment of probabilities to sentences. To appreciate the difference between this and the expression of statistical information consider the two assertions: "The probability that Tweety can fly is 0.75," and "More than 75% of all birds can fly." The first statement is an expression of a degree of belief. It is expressing the internal state of some agent—an agent who believes the assertion "Tweety can fly" to degree 0.75. It is not an objective assertion about the state of the world (i.e., an assertion that is independent of any believers). In the world Tweety can either fly or not fly—there is no probability involved. The second statement, on the other hand, is making an objective true-false assertion about the state of the world; i.e., in the world there is some percentage of birds that can fly and this percentage is either 75% or some other number.[1]

This example shows that there is an essential difference between the attachment of a probability to a sentence and the expression of a statistical assertion. As has been demonstrated in work presented at the previous uncertainty workshop (Bacchus [7]) probabilities attached to sentences, which have been the focus of previous work on probability logics, *op. cit.*, are not capable of efficiently expressing statistical assertions.

There has been some work similar to **Lp**. This work is discussed in more detail in Bacchus [8].

## 3 Types of Statistical Knowledge

Statistical information can be categorized into many different types. The development of **Lp** was guided by a desire to represent as many different types of statistical knowledge as possible. The key consideration was the desire to represent qualitative statistical knowledge, i.e., not only the types of statistical knowledge used in statistics but also the types of "commonsense" statistical knowledge that would be useful in AI domains. The following is an incomplete list of some different types of statistical information that **Lp** is capable of expressing.

---

[*]Support for preparing this paper was provided through a grant from the University of Waterloo, and NSERC grant OGP0041848. Parts of this work have been previously reported at CSCSI-88 [19].

[1] As stated it is clear that it is extremely unlikely that the actual percentage is exactly 75%. More likely that it is in some interval around 75%. **Lp** is also capable of making such interval assertions.

**Relative:** Statistical information may be strictly comparative, e.g., the assertion "More politicians are lawyers than engineers."

**Interval:** We may know that the proportion is in a certain range, e.g., the assertion "Between 75% to 99% of all politicians are lawyers."

**Functional:** We may know that a certain statistic is functionally dependent on some other measurement, e.g., "The proportion of flying birds decreases as weight increases." This type of functional dependence in an uncertainty measure is prominent in the medical domain.

**Independence:** We may know that two properties are statistically independent of each other. Work by Pearl and his associates has demonstrated the importance of this kind of knowledge ([9, 10, 11]).

## 4 Syntax and Semantics

**Lp** is based on two fairly straightforward ideas. First, there is a probability distribution over the domain of discourse. This means that any set of domain individuals can be assigned a probability. Through the use of open formulas (i.e., formulas with free variables) we can assert that various sets of domain individuals possess certain probabilities. An open formula can be viewed, as in lambda abstraction, as specifying a set of domain individuals—the set of individuals which satisfy that formula. For example the open formula "$\text{Bird}(x)$" can be viewed as denoting the set of birds, i.e., the set of individuals that satisfy the formula. Sentences in **Lp** can be used to assert that the probability of this set (i.e., the measure of the set of individuals that satisfy the formula) possesses various properties. For example, the **Lp** sentence "$[\text{Bird}(x)]_x > 0.9$" asserts that the probability of the set of birds has the property that it is greater than 0.9.[2]

The second idea is to have a field of numbers in the semantics as a separate sort. With numbers as a separate sort the probabilities become individuals in the logics. That is, the probabilities become numeric terms[3] and, by asserting that these terms stand in various numeric relationships with other terms, we can assert various qualitative relationships between these probabilities. In the above example, '$[\text{Bird}(x)]_x$' is a numeric term, and the sentence asserts that it stands in the 'greater-than' relation with the numeric term '0.9'. The existence of numbers as a separate sort also allows the use of 'measuring' functions, functions that map individuals to numbers. An example of such a function is 'Weight', which maps individuals to a number representing their weight (in some convenient units). The measuring functions greatly increase the expressiveness of the logic.[4]

## 5 Syntax

We now present in a bit more detail the exact syntax of **Lp**. This description should give the reader a better idea of the types of sentences that one can form in the language.

We start with a set of constant, variable, function, and predicate symbols. The constants, variables, and predicates can be of two types, either field or object.[5] The function symbols come in three different types: object, field, and measuring functions. The measuring functions will usually have special names like Weight or Size.

Along with these symbols we also have a set of distinguished symbols, including the following field symbols: 1, 0 (constants), $=$, $\geq$ (predicates), $+$, $-$, $\times$, and $\div$[6] (functions). The symbol $=$ is also used to represent the object equality predicate. Also included is the logical connective '$\wedge$', the quantifier '$\forall$', and the probability term formers '[', ']'.

### 5.1 Formulas

The major difference between the formulas of **Lp** and the formulas of first order logic is the manner in which terms are built up.

**T0)** A single object variable or constant is an *o-term*; a single field variable or constant is an *f-term*.

**T1)** If $f$ is an n-ary object (field) function symbol and $t_1, \ldots, t_n$ are o-terms (f-terms) then $f(t_1 \ldots t_n)$ is an *o-term* (*f-term*). If $\nu$ is an n-ary measuring function symbol and $t_1, \ldots, t_n$ are o-terms then $\nu(t_1 \ldots t_n)$ is an *f-term*.

**T2)** If $\alpha$ is a formula and $\vec{x}$ is a vector of $n$ object variables, $\langle x_1, \ldots, x_n \rangle$, then $[\alpha]_{\vec{x}}$ is an *f-term*.[7]

The formulas of **Lp** are built up in the standard manner, with the added constraint that predicates can only apply to terms of the same type. The notable difference with first order logic is that *f-terms* can be generated from formulas by the probability term former. For example, from the formula "Have(y,x)∧Zoo(x)" the f-term "[Have(y,x)∧Zoo(x)]x" can be generated. This term can then be used to generate new formulas of arbitrary complexity, e.g.,

$$(\forall yz)\Big(\text{Rare}(y) \land \neg\text{Rare}(z) \land \text{Animal}(y) \land \text{Animal}(z)$$
$$\rightarrow [\text{Have}(z,x) \land \text{Zoo}(x)]_x > [\text{Have}(y,x) \land \text{Zoo}(x)]_x\Big)$$

---

[2]This unconditional probability does not make much sense; it is through the use of conditional probabilities that meaningful statistical assertions can be made. For example, the **Lp** sentence "$[\text{Fly}(x)|\text{Bird}(x)]_x > 0.9$" makes an assertion about the relative probability of flying birds among birds, i.e., about the proportion of birds that fly.

[3]This means that the probabilities are field-valued not real-valued. There are technical difficulties with using the reals instead of a field of numbers. In particular, it is not possible to give a complete axiomatization of the reals without severely restricting the expressiveness of the logic. We can be assured, however, that the field of numbers will always contain the rational numbers, so the probabilities can be any rational number that we wish (in the range 0–1, of course).

[4]These "measuring" functions are called random variables in statistics, but I avoid that terminology to eliminate possible confusion with the ordinary variables of **Lp**.

[5]When there is a danger of confusion the field symbols will be written in a **bold** font.

[6]The division function is added by extending the language through definition. See [8] for the technical details.

[7]Note, $\vec{x}$ does not have to include all of the free variables of $\alpha$. If it does not we have a term with free variables which must be bound by other quantifiers or probability term formers to produce a sentence.

In this formula some of the variables are universally quantified while the 'x' is bound by the probability term former. The intuitive content of this formula can be stated as follows: if there are two animals one of which is rare while the other is not then the measure (probability) of the set of zoos which have the rare animal is less than the measure of the set of zoos which have the non-rare animal.

Through standard definitions we add $\lor$, $\rightarrow$, $\exists$, and the extended set of field inequality predicates, $\leq$, $<$, $>$, and $\in$ (denoting membership in an interval). We use infix form for the predicate symbols $=$ and $\geq$ as well as for the function symbols $+$, $\times$, $-$, and $\div$.

Conditional probabilities are represented in **Lp** with the following abbreviation.

**Definition 1**

$$[\alpha|\beta]_{\vec{x}} =_{df} [\alpha \land \beta]_{\vec{x}} \div [\beta]_{\vec{x}},$$

### 5.2 Semantic Model

This section outlines the semantic structure over which **Lp** is interpreted. As indicated above it consists of a two sorted domain (individuals and numbers) and a probability distribution over the set of individuals. What was not discussed was the need for a distribution over all vectors of individuals. This is necessary since the open formulas used to generate the probability terms may have more than one free variable. Hence one may need to examine the probability of the set of vectors of individuals which satisfy a given formula.

An **Lp**-Structure is defined to be the tuple $\mathcal{M}$:

$$\langle (\mathcal{O}, R_{\mathcal{O}}, F_{\mathcal{O}}), (\mathcal{F}, R_{\mathcal{F}}, F_{\mathcal{F}}), \Psi, \{\mu_n \mid n = 1, 2 \ldots\}\rangle$$

Where:

a) $(\mathcal{O}, R_{\mathcal{O}}, F_{\mathcal{O}})$: $\mathcal{O}$ represents a finite set of individual objects,[8] $R_{\mathcal{O}}$ a set of relations, and $F_{\mathcal{O}}$ a set of functions, both of any arity.

b) $(\mathcal{F}, R_{\mathcal{F}}, F_{\mathcal{F}})$: $\mathcal{F}$ represents a totally ordered field of numbers along with a set of relations, $R_{\mathcal{F}}$, and functions, $F_{\mathcal{F}}$, on the field.

c) $\Psi$ represents a set of measuring functions, functions from $\mathcal{O}^n$ to $\mathcal{F}$.

d) $\{\mu_n \mid n = 1, 2, \ldots\}$ is a sequence of probability functions. Each $\mu_n$ is a set function whose domain includes the subsets of $\mathcal{O}^n$ defined by the formulas of **Lp**,[9] whose range is $\mathcal{F}$, and which satisfies the axioms of a probability function (i.e., $\mu_n(A) > 0$, $\mu_n(A \cup B) = \mu_n(A) + \mu_n(B)$ if $A \cap B = \emptyset$, and $\mu_n(\mathcal{O}^n) = 1$).

The sequence of probability functions is a sequence of product measures. That is, for any two sets $A \in \mathcal{O}^n$ and $B \in \mathcal{O}^m$ and their Cartesian product $A \times B \in \mathcal{O}^{n+m}$, if $A \in domain(\mu_n)$ and $B \in domain(\mu_m)$, then $A \times B \in domain(\mu_{n+m})$ and $\mu_{n+m}(A \times B) = \mu_n(A) \times \mu_m(B)$.

The product measure ensures that the probability terms satisfy certain conditions of coherence. For example, the order of the variables cited in the probability terms makes no difference, e.g., $[\alpha]_{x,y} = [\alpha]_{y,x}$. Another example is that the probability terms are unaffected by tautologies, e.g., $[P(x) \land (R(y) \lor \neg R(y))]_{(x,y)} = [P(x)]_x$.

It should be noted that this constraint on the probability functions is not equivalent to a restrictive assumption of independence, sometimes found in probabilistic inference engines (e.g., the independence assumptions of the Prospector system [12], see Johnson [13]).

### 5.3 Semantics of Formulas

The formulas of **Lp** are interpreted with respect to the semantic structure in the same manner as first order formulas are interpreted with respect to first order structures. The only difference is that we have to provide an interpretation of the probability terms. As indicated above the probability terms denote the measure (probability) of the set of satisfying instances of the formula. In more detail:

We define a correspondence, called an interpretation, between the formulas and the **Lp**-Structure $\mathcal{M}$ augmented by the truth values $\top$ and $\bot$ (true and false). An interpretation maps all of the symbols to appropriate entities in the **Lp**-Structure, including giving an initial assignment to all of the variables.

These assignments serve as the inductive basis for an interpretation of the formulas. This interpretation is built up in the same way as in first order logic, with the added consideration that universally quantified object variables range over $\mathcal{O}$ while universally quantified field variables range over $\mathcal{F}$. The only thing which needs to be demonstrated is the semantic interpretation of the probability terms.

Let $\sigma$ be an interpretation of **Lp**. Let $\sigma(\vec{x}/\vec{a})$, where $\vec{a} = \langle a_1, \ldots, a_n\rangle$ and $\vec{x} = \langle x_1, \ldots, x_n\rangle$ are vectors of individuals and variables (of matching type), denote a new interpretation identical to $\sigma$ except that $(x_i)^{\sigma(\vec{x}/\vec{a})} = a_i$, $(i = 1, \ldots, n)$.

The probability terms are given the following semantic interpretation: For the f-term $[\alpha]_{\vec{x}}$,

$$([\alpha]_{\vec{x}})^\sigma = \mu_n\{\vec{a}|\alpha^{\sigma(\vec{x}/\vec{a})} = \top\}.$$

In other words, the probability term denotes the probability of the set of satisfying instances of the formula. Since $\mu_n$ is a probability function which maps to the field of numbers $\mathcal{F}$, it is clear that $[\alpha]_{\vec{x}}$ denotes an element of $\mathcal{F}$ under the interpretation $\sigma$; thus, it is a valid f-term.

## 6 Examples of Representation

We can now give a indication of the representational power of **Lp**. By considering the semantic interpreta-tion of the formulas it should be reasonably clear that the formulas do in fact represent the gist of the stated English assertions.[10]

---
[8] We restrict ourselves to finite domains to avoid the difficulty of sigma additivity. This issue is dealt with in [8].

[9] This set of subsets can be shown to be a field of subsets [8].

1. *More politicians are lawyers than engineers.*

   $[\text{Lawyer}(x)|\text{Politician}(x)]_x$
   $> [\text{Engineer}(x)|\text{Politician}(x)]_x.$

2. *The proportion of flying birds decreases with weight.* Here y is a field variable.

   $\forall y ([\text{fly}(x)|\text{bird}(x) \wedge \text{weight}(x) < y]_x$
   $> [\text{fly}(x)|\text{bird}(x) \wedge \text{weight}(x) > y]_x).$

3. *Given R the property P is independent of Q.* This is the canonical tri-functional expression of independence (see Pearl [9]).

   $[P(x) \wedge Q(x)|R(x)]_x = [P(x)|R(x)]_x \times [Q(x)|R(x)]_x.$

   Thus **Lp** can represent finely grained notions of independence at the object language level.

4. Quantitative notions from statistics, e.g, *The height of adult male humans is normally distributed with mean 177cm and standard deviation 13cm*:

   $\forall yz ([\text{height}(x) \in (y,z)|\text{Adult\_male}(x)]_x$
   $= \text{normal}(y,z,177,13)).$

   Here normal is a field function which, given an interval $(y,z)^{11}$, a mean, and a standard deviation, returns the rational number approximation[12] of the integral of a normal distribution, with specified mean and standard deviation, over the given interval.

## 7 Deductive Proof Theory

This section outlines the deductive proof theory of **Lp**. The proof theory provides a specification for wide class of valid inferences that can be made from a body of knowledge expressed in **Lp**. In particular, it provides a full specification for most probabilistic inferences, including Baysian inference, all first order inferences, as well as inferences which follow from the combination of qualitative and quantitative as well as statistical and logical knowledge.

The proof theory consists of a set of axioms and rules of inference, and can be shown to be both sound. It can also be shown to be complete with respect to various classes of models. The proof theory for **Lp** is similar to the proof theory for ordinary first order logic. The major change is that two new sets of axioms must be introduced, one to deal with the logic of the probability function, and another set to define the logic of the field $\mathcal{F}$.

The axioms include the axioms of first order logic (e.g., [15]) along with the axioms of a totally ordered field (MacLane [14]). There are also various axioms which specify the behavior of the probability terms. We give some examples of these axioms to give a indication of their form.

**Some of the Probability Function Axioms**

**P1)** $\forall x_1 \ldots \forall x_n \alpha \rightarrow [\alpha]_{\vec{x}} = 1$,
where $\vec{x} = \langle x_1, \ldots, x_n \rangle$, and every $x_i$ is an object variable.

**P2)** $[\alpha]_{\vec{x}} \geq 0.$

**P3)** $[\alpha]_{\vec{x}} + [\neg \alpha]_{\vec{x}} = 1.$

**P4)** $[\alpha]_{\vec{x}} + [\beta]_{\vec{x}} \geq [\alpha \vee \beta]_{\vec{x}}.$

**P5)** $[\alpha \wedge \beta]_{\vec{x}} = 0 \rightarrow [\alpha]_{\vec{x}} + [\beta]_{\vec{x}} = [\alpha \vee \beta]_{\vec{x}}.$

The first axiom simply says that if all individuals satisfy a given formula then the probability of this set is one (i.e., the probability summed over the entire domain is one). The other axioms state similar facts from the calculus of probabilities.

**Rule of inference**

The only rule of inference is *modus ponens*, i.e., from $\{\alpha, \alpha \rightarrow \beta\}$ infer $\beta$.

If we also have an axiom of finiteness (see Halpern [5]) then the above axioms and rule of inference comprise a sound and complete proof theory for the class of models we have defined here (i.e., models in which $\mathcal{O}$ is bounded in size and where the probabilities are field valued). Let $\Phi$ be a set of Lp sentences. We have:

**Theorem 2 (Completeness)** *If $\Phi \models \alpha$, then $\Phi \vdash \alpha$.*

**Theorem 3 (Soundness)** *If $\Phi \vdash \alpha$, then $\Phi \models \alpha$.*

**Lemma 1** *The following are provable[13] in* **Lp***:*

a) $([\alpha \rightarrow \beta]_{\vec{x}} = 1 \wedge [\beta \rightarrow \alpha]_{\vec{x}} = 1) \rightarrow [\alpha]_{\vec{x}} = [\beta]_{\vec{x}}.$

b) $[\alpha \vee \beta]_{\vec{x}} = [\alpha]_{\vec{x}} + [\beta]_{\vec{x}} - [\alpha \wedge \beta]_{\vec{x}}.$

The following gives an indication of the scope of the proof theory.

**Example 1** *Nilsson's Probabilistic Entailment*

Nilsson [1] shows how the probabilities of sentences in a logic are constrained by known probabilities, i.e., constrained by the probabilities of a base set of sentences. For example, if $[P \wedge Q] = 0.5$, then the values of $[P]$ and $[Q]$ are both constrained to be $\geq 0.5$. Nilsson demonstrates how the implied constraints of a base set of sentences can be represented in a canonical manner, as a set of linear equations. These linear equations can be used to identify the strongest constraints on the probability of a new sentence, i.e., the tightest bounds on its probability. These constraints are, in Nilsson's terms, probabilistic entailments.

---

[10] It should be noted that the aim is to give some illustrative examples, not to capture all of the nuances of the English assertions.

[11] One would probably want to constrain the values of y and z further, for example, y < z.

[12] A rational number approximation is returned since the numbers are from a totally ordered field, not necessarily the reals. It is well known that every totally ordered field contains the rationals (MacLane [14]).

[13] That is, deducible directly from the axioms.

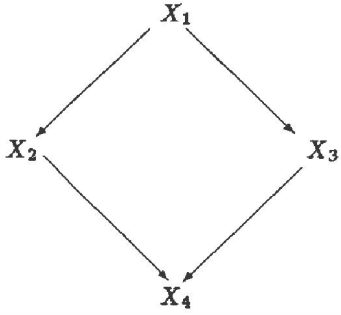

Figure 1: A Bayes' Net

These bounds are simply consequences of the laws of probability. And, since the proof theory of **Lp** is complete with respect to the finite sample spaces that Nilsson uses, these probabilistic entailments can be deduced from the proof theory of **Lp**.

For example, if the base set in Nilsson's logic is $\{[P]=0.6, [P \to Q]=0.8\}$, probabilistic entailment gives the conclusion $0.4 \leq [Q] \leq 0.8$. If we write the symbols $P$ and $Q$ as one place predicates, then in **Lp** the knowledge could be represented by the following set: $\{[P(x)]_x = 0.6, [P(x) \to Q(x)]_x = 0.8\}$.

From this knowledge it is easy to deduce the bounds $[0.4, 0.8]$ on the probability term $[Q(x)]_x$.

**Example 2** *Bayesian Networks.*

Bayes' theorem is immediate in **Lp**.

**Lemma 2 (Bayes' Theorem)** *The following is provable in* **Lp**:
$$[\beta|\alpha]_{\vec{x}} = [\alpha|\beta]_{\vec{x}} \times \frac{[\beta]_{\vec{x}}}{[\alpha]_{\vec{x}}}.$$

Consider the Bayes' Net in figure 1. If all of the variables $X_1$–$X_4$ are propositional (binary) variables one could write them as one place predicates in **Lp**. Hence, the open formula '$X_1(x)$', for example, would denote the set of individuals with property $X_1$. The Bayes' Net gives a graphical device for specifying a product form for the joint distribution of the variables $X_i$ [6]. In this case the distribution represented by the Bayes' Net in the figure could also be specified by the **Lp** sentence

$$\begin{aligned}
&[X_1(x) \wedge X_2(x) \wedge X_3(x) \wedge X_4(x)]_x \\
&= [X_4(x)|X_3(x) \wedge X_2(x)]_x \times [X_3(x)|X_1(x)]_x \\
&\quad \times [X_2(x)|X_1(x)]_x \times [X_1(x)]_x
\end{aligned}$$

It can easily be demonstrated that any probability distribution which satisfies this equation will also satisfy every equation of the same form with any number of the predicates negated (uniformly). For example, the equation

$$\begin{aligned}
&[X_1(x) \wedge \neg X_2(x) \wedge X_3(x) \wedge \neg X_4(x)]_x \\
&= [\neg X_4(x)|X_3(x) \wedge \neg X_2(x)]_x \times [X_3(x)|X_1(x)]_x \\
&\quad \times [\neg X_2(x)|X_1(x)]_x \times [X_1(x)]_x
\end{aligned}$$

will be satisfied by every probability distribution which satisfies the first equation. Furthermore, the proof depends only on finite properties of the probability function, i.e., only on properties true of the field valued probabilities used in the **Lp**-structure. Hence, by the completeness result, all such equations will be provable from the first via **Lp**'s proof theory.

This means that the behavior of the Bayes' net is captured by the first **Lp** sentence. That is, the fact that this product decomposition holds for every instantiation of the propositional variables $X_i$ is captured by the proof theory.

In addition to the structural decomposition Bayes' nets must provide a quantification of the links. This means the conditional probabilities in the product must be specified. In this example if we add the **Lp** sentences $\{[X_1(x)]_x = 0.5, [X_2(x)|X_1(x)]_x = .75, [X_3(x)|X_1(x)]_x = .4, [X_4(x)|X_2(x) \wedge X_3(x)]_x = .3\}$, we can then determine the probabilities of any of the individual variables given an instantiation of some of the other variables, e.g., the values of terms like $[X_1(x)|X_2(x) \wedge \neg X_4(x)]_x$. Again these probabilities will be semantically entailed by the product decomposition and by the link conditional probabilities. Thus, the new probability values will be provable from the proof theory.

Of course the proof theory has none of the computational advantages of the Bayes' net. However, what is important is that **Lp** gives a declarative representation of the net. The structure embedded in the net is represented in a form that can be reasoned with and can be easily changed. There is also the possibility of automatically compiling Bayes' net structures from declarative **Lp** sentences. Furthermore, the proof theory captures all of the Baysian reasoning within its specification, and offers the possibility of integrating Bayes' net reasoning with more general logical and qualitative statistical reasoning. Hence the proof theory gives unifying formalism in which both types of inferences could be understood.

## 8 Degrees of Belief

Besides their use in expressing statistical information, probabilities have an important use in expressing degrees of belief. One can assert that PROB[Fly(Tweety)] > .75, indicating that one's degree of belief in the assertion Fly(Tweety) is greater than 0.75. Interestingly, **Lp** is not capable of expressing such probabilities. It can be shown that the probability of any sentence (i.e., formula with no free variables) is either 1 or 0 in **Lp**. This fact is interesting because, as was demonstrated in [7], probability logics capable of assigning probabilities to sentences cannot (easily) represent statistical probabilities. Hence, these two types of probability logics have very different uses which coincide with their very different semantics.

However, one advantage of a logic like **Lp** is that it can be used to generate statistically founded degrees of belief, via a system of direct inference (Kyburg [16], Pollock [17]). Degree of belief probabilities generated in this manner have a number of advantages over purely subjective probabilities [18]; not the least of which is that they yield degrees of belief which are founded on empirical experience.

A preliminary system of statistically founded degrees of belief is presented in [19]. Further work in this area is in progress and will be reported in [8]. The following example should serve to illustrate the basic idea behind this system.

**Example 3** *Belief Formation*

Say we know that we have the following Lp knowledge base

$$KB = \begin{array}{l} [\text{Fly}(x)|\text{Bird}(x)]_x > 0.9 \\ \text{Bird(Tweety)} \end{array}$$

That is, we know that more than 90% of all birds fly, and that Tweety is a bird. Say that we want to generate a degree of belief about Fly(Tweety), i.e., Tweety's flying ability. We can accomplish this by considering what is know about Tweety (i.e., what is provable from our knowledge base), and then equating our degree of belief with the statistical probability term which results when we substitute a variable for the constant Tweety. This yields

$$\text{PROB}(\text{Fly(Tweety)}|\text{Bird(Tweety)}) = [\text{Fly}(x)|\text{Bird}(x)]_x,$$

which by our knowledge base is greater than 0.9. Semantically, this can be interpreted in the following manner: our degree of belief that Tweety can fly, given that all we know about Tweety is that he is a bird, is equal to the proportion of birds that can fly. The main complexities arise when we know other things about Tweety, e.g., when we know that Tweety is a penguin as well as a bird.

## Acknowledgments

Thanks to the referees for some helpful criticisms on my submitted draft.